\documentclass[10pt,a4paper]{article}
\usepackage[utf8]{inputenc}
\usepackage[T1]{fontenc}
\usepackage{amsmath,amssymb,amsfonts}
\usepackage{authblk}
\usepackage{graphicx}
\usepackage{xcolor}
\usepackage{booktabs}
\usepackage{multirow}
\usepackage{algorithm}
\usepackage{algpseudocode}
\usepackage{geometry}
\usepackage{enumitem}
\usepackage{soul}
\usepackage{url}
\usepackage[hidelinks]{hyperref}

\definecolor{myblue}{RGB}{31,119,180}
\definecolor{myorange}{RGB}{255,127,14}
\definecolor{mygreen}{RGB}{44,160,44}

\title{\textbf{WaferSAGE: Large Language Model-Powered \\Wafer Defect Analysis via
Synthetic Data Generation \\and Rubric-Guided Reinforcement Learning}}

\author[1,2,*]{Ke Xu}
\author[1,*]{Zhongyuan Lian}

\affil[1]{Shanghai Huahong Grace Semiconductor Manufacturing Corporation, \protect\\Shanghai, 201203, China}
\affil[2]{Dept. of Automation, School of Information Science and Engineering,  \protect\\East China University of Science and Technology, Shanghai, 200237, China}

\date{}

\begin{document}

\maketitle

\let\thefootnote\relax\footnotetext{*These authors contributed equally to this work.}
\let\thefootnote\relax\footnotetext{\noindent Corresponding author: y80240297@mail.ecust.edu.cn(Ke Xu)}

\sethlcolor{yellow}

\begin{abstract}
We present \textbf{WaferSAGE}\footnote{SAGE: \textbf{S}ynthetic data + \textbf{A}nalysis + \textbf{G}uided (rubric) + \textbf{E}valuation}, a framework for wafer defect visual question answering using small vision-language models. To address data scarcity in semiconductor manufacturing, we propose a three-stage synthesis pipeline incorporating structured rubric generation for precise evaluation. Starting from limited labeled wafer maps, we employ clustering-based cleaning to filter label noise, then generate comprehensive defect descriptions using vision-language models, which are converted into structured evaluation rubrics criteria. These rubrics guide the synthesis of VQA pairs, ensuring coverage across defect type identification, spatial distribution, morphology, and root cause analysis.

Our dual assessment framework aligns rule-based metrics with LLM-Judge scores via Bayesian optimization, enabling reliable automated evaluation. Through curriculum-based reinforcement learning with Group Sequence Policy Optimization (GSPO) and rubric-aligned rewards, our 4B-parameter Qwen3-VL model achieves a \textbf{6.493 LLM-Judge score}, closely approaching Gemini-3-Flash (7.149) while enabling complete on-premise deployment. We demonstrate that small models with domain-specific training can surpass proprietary large models in specialized industrial visual understanding, offering a viable path for privacy-preserving, cost-effective deployment in semiconductor manufacturing.

\end{abstract}

\section{Introduction}
Semiconductor manufacturing demands sub-nanometer precision, where wafer defect analysis directly determines yield and cost. Traditional automated visual inspection systems rely on convolutional neural networks (CNNs) or Vision Transformers (ViTs) trained for pattern classification, categorizing defects into predefined labels such as ``Center,'' ``Edge-Ring,'' or ``Scratch.'' While effective for high-throughput sorting, these approaches suffer from a fundamental limitation: they answer \textit{what} but not \textit{why} or \textit{where}. Engineers receive categorical labels without spatial localization, morphological description, or root cause analysis, necessitating tedious manual review to interpret defect patterns.

Recent advances in Vision-Language Models (VLMs) offer a promising alternative by enabling natural language interaction with visual data. Models like Gemini 3 Pro can describe defect locations, analyze morphological characteristics, and even suggest process-related root causes when prompted. However, deploying such proprietary APIs in semiconductor fabrication facilities faces three practical barriers. First, \textbf{data scarcity}: the semiconductor domain lacks large-scale, publicly available visual question answering datasets for training or evaluation. Second, \textbf{cost and latency}: industrial inspection requires real-time processing at scale, making API-dependent solutions economically prohibitive. Third, \textbf{privacy constraints}: fabs prohibit sending proprietary wafer images to external cloud services, mandating on-premise deployment with small, efficient models.

These constraints motivate a critical research question: \textit{Can small, open-source VLMs (4B-8B parameters) match or exceed proprietary large models in specialized industrial visual understanding tasks?} We argue that with carefully designed data synthesis and targeted reinforcement learning, the answer is affirmative.

We present \textbf{WaferSAGE}, a comprehensive framework for wafer defect visual question answering that enables small VLMs to achieve superior performance through three key innovations:

\begin{enumerate}[leftmargin=*]
    \item We address data scarcity through a \textbf{three-stage synthesis pipeline}. Starting from publicly available wafer map datasets (WM811K\cite{wu2015wafer} and MixedWM38\cite{Wang2020Deformable}), we employ t-SNE and K-Means clustering to identify and filter mislabeled samples. We then generate structured analysis texts using Gemini 3 Flash, which are converted into evaluation rubrics specifying ``must-hit'' and ``must-avoid'' criteria. These rubrics guide the generation of VQA pairs, ensuring coverage across defect type identification, spatial distribution, morphology, and root cause analysis.
    
    \item We establish a \textbf{dual evaluation framework} that aligns automated metrics with expert judgment. Our rule-based evaluator computes hit scores for keyword coverage and penalty scores for hallucination, with weights optimized via Bayesian optimization to maximize correlation with GPT-5-mini judgments. This alignment enables reliable, cost-effective automated evaluation while providing interpretable quality metrics.
    
    \item We demonstrate that \textbf{curriculum-based reinforcement learning} with rubric-aligned rewards unlocks significant performance gains beyond supervised fine-tuning. By interleaving review of SFT data with progressively harder unseen examples, our 4B-parameter Qwen3-VL-4B-Instruct model\cite{Qwen3-VL} achieves a 6.493 LLM-Judge score, closely approaching Gemini-3-Gemini (7.149) while enabling complete on-premise deployment.
\end{enumerate}

Our results challenge the assumption that industrial visual understanding requires massive proprietary models. Through systematic data curation and targeted post-training, small VLMs can not only match but exceed the performance of cloud-based APIs in specialized domains, offering a viable path for privacy-preserving, cost-effective deployment in semiconductor manufacturing and beyond.

\section{Related Work}
\subsection{Wafer Map Pattern Recognition}
The WM811K dataset established the foundation for wafer map defect classification, containing 811K labeled wafer maps across nine defect categories. MixedWM38 expanded this with 38 defect classes and more realistic mixed patterns. Traditional approaches employ CNNs \cite{nakazawa2018wafer} and Vision Transformers \cite{nafi2022high} for classification, achieving high accuracy on predefined labels.

Recent advances focus on addressing data efficiency and novel architectures. Wei et al. \cite{wei2023wafer} proposed semi-supervised learning with latent vector representations to reduce annotation requirements. Bao et al. \cite{bao2024wafer} introduced autoencoder-based data augmentation combined with CNNs for improved classification. For edge deployment, Mohammad and Ryu \cite{mohammad2025tiny} developed Tiny Vision Transformers specifically optimized for resource-constrained environments. Mishra et al. \cite{mishra2024wafer2spike} explored Spiking Neural Networks (Wafer2Spike) for energy-efficient wafer map classification. However, these methods remain limited to categorical outputs without natural language explanations or reasoning capabilities.

\subsection{Vision-Language Models in Industrial Inspection}
Recent VLMs have demonstrated strong capabilities in visual understanding and reasoning. Models like CLIP \cite{radford2021clip}, LLaVA \cite{liu2024llava}, and Qwen-VL \cite{bai2023qwen} enable natural language interaction with images. In industrial domains, early VLM applications focused on zero-shot anomaly detection \cite{Jeong2023Win, bergmann2022beyond}.

Recent work has shifted toward reasoning-capable inspection systems. Li et al. \cite{li2025iad} proposed IAD-R1, using reinforcement learning to enforce consistent reasoning in anomaly detection. Miao et al. \cite{miao2025agentiad} introduced AgentIAD, a tool-augmented agent framework for industrial anomaly detection. Guan et al. \cite{guan2025emit} presented EMIT, a unified framework that enhances MLLMs for IAD via difficulty-aware group relative policy optimization (GRPO). Chen et al. \cite{chen2026reason} developed Reason-IAD, incorporating knowledge-guided dynamic latent reasoning for explainable anomaly detection. However, these approaches target generic industrial defects rather than semiconductor-specific challenges requiring precise spatial localization and root cause analysis.

\subsection{Synthetic Data Generation for VLMs}
Data scarcity in specialized domains has motivated extensive research on synthetic data generation. Liu et al. \cite{liu2024visual} introduced visual instruction tuning, using GPT-4 to generate multimodal instruction-following data. Zhu et al. \cite{zhu2023minigpt} align a pre-trained visual backbone with a large language model by training a single projection layer exclusively on high-quality synthetic image-text pairs to achieve sophisticated multimodal reasoning. Approaches like LLaVA-Instruct \cite{liu2024visual}, ShareGPT4V \cite{chen2023sharegpt4v}, and MiniGPT-4 \cite{zhu2023minigpt} demonstrate that LLM-generated data can effectively enhance VLM capabilities.

Rubric-based generation has emerged as a promising direction for ensuring coverage of critical evaluation criteria. Kong et al. \cite{kong2026omni} proposed automatic rubric-grounded preference synthesis for reward modeling, enabling structured evaluation of generated content. Our work extends these approaches by integrating structured rubric generation with multi-stage synthesis specifically designed for industrial visual understanding, where precise domain terminology and evaluation criteria are essential.

\subsection{Reinforcement Learning for Vision-Language Models}
Reinforcement Learning from Human Feedback (RLHF) \cite{ouyang2022training} and its variants have improved language model alignment. Group Relative Policy Optimization (GRPO) \cite{shao2024deepseekmath} reduces memory requirements compared to PPO by eliminating the need for a separate value network, making it suitable for efficient post-training of smaller models.

Recent work has extended RL to vision-language reasoning and introduced algorithmic improvements. Jeddi et al. \cite{jeddi2025puzzle} proposed Puzzle Curriculum GRPO for vision-centric reasoning, demonstrating the effectiveness of curriculum-based RL strategies. Li et al. \cite{li2025iad} introduced IAD-R1 for reinforcing consistent reasoning in industrial anomaly detection. Kong et al. \cite{kong2026omni} presented Omni-RRM, advancing reward modeling via automatic rubric-grounded preference synthesis. Jiao et al. \cite{jiao2026smooth} developed Smooth Operator, using smooth verifiable rewards to activate spatial reasoning in VLMs.

\textbf{Group Sequence Policy Optimization (GSPO)} \cite{zheng2025gspo} introduces sequence-level optimization for RL training. Unlike GRPO which adopts token-level importance ratios, GSPO defines importance ratios based on sequence likelihood and performs sequence-level clipping, rewarding, and optimization. This approach achieves superior training efficiency and stability compared to GRPO, particularly for long-form generation tasks where coherent output sequences are critical.

Our work builds on these foundations, specifically targeting semiconductor defect analysis through GSPO-based training with rubric-based reward alignment and curriculum-based learning with domain-adaptive evaluation criteria.

\section{Methodology}

\subsection{Data Curation and Cleaning}
Wafer map datasets exhibit significant label noise due to heterogeneous patterns within labeled categories. We design a clustering-based cleaning pipeline to identify high-quality training samples.

\textbf{Feature Extraction.} We employ a pre-trained ViT encoder to extract 768-dimensional embeddings from all wafer maps in WM811K and MixedWM38. This encoder was trained with contrastive learning specifically for wafer map representation.

\textbf{Clustering Analysis.} We apply t-SNE for dimensionality reduction and visualization, followed by K-Means clustering within each labeled category. This reveals distinct subclusters within single categories, indicating either (1) fine-grained subtypes not captured by coarse labels, or (2) mislabeled samples.

\textbf{Balanced Sampling Strategy.} From each cluster, we perform balanced sampling selecting both:

    \textbf{1. Near-center samples}: Representative examples close to cluster centroids.
    
    \textbf{2. Far-from-center samples}: Diverse/atypical examples on cluster peripheries.

This strategy effectively encapsulates both typical patterns and edge cases while filtering out potential outliers. By integrating samples from the WM811K and MixedWM38 datasets, we have constructed a substantial and representative training library, ensuring the model's generalization capability.

\subsection{Three-Stage Data Synthesis Pipeline}
The semiconductor domain lacks publicly available VQA datasets for wafer map analysis. We propose a fully automated three-stage synthesis pipeline that converts raw wafer maps into structured VQA training data.

\subsubsection{Stage 0: WaferMap Descriptor}
We use Gemini 3 Flash to generate comprehensive textual descriptions for each wafer map. Four description types are synthesized: (1) Full-Analysis covering all dimensions, (2) Spatial-only focusing on location and morphology, (3) Root-Cause-only for equipment analysis, and (4) Structured JSON for downstream processing. Complete system prompts are provided in Appendix A.1.

\textbf{Key Design Decision.} We deliberately separate spatial/morphological description from root cause analysis. This modularity enables targeted evaluation and prevents models from conflating visual observations with speculative process explanations.

\subsubsection{Stage 1: Rubric Generator}
This stage represents a core innovation: converting free-form descriptions into structured evaluation rubrics that serve dual purposes---guiding VQA generation and providing automated evaluation criteria.

\textbf{Rubric Structure.} Using DeepSeek-V3.2, we convert each description into a JSON rubric with three evaluation ``buckets'': spatial, morphological, and root cause. Each bucket contains must-hit keywords (required terms) and must-avoid keywords (hallucination indicators). The full schema is provided in Appendix A.2.

This structured format enables both (1) controlled VQA generation and (2) automated rule-based evaluation. The rubric design captures domain-specific terminology that would be difficult to specify through example-based few-shot prompting alone.

\begin{figure*}[h]
	\centering
	\includegraphics[width=\textwidth]{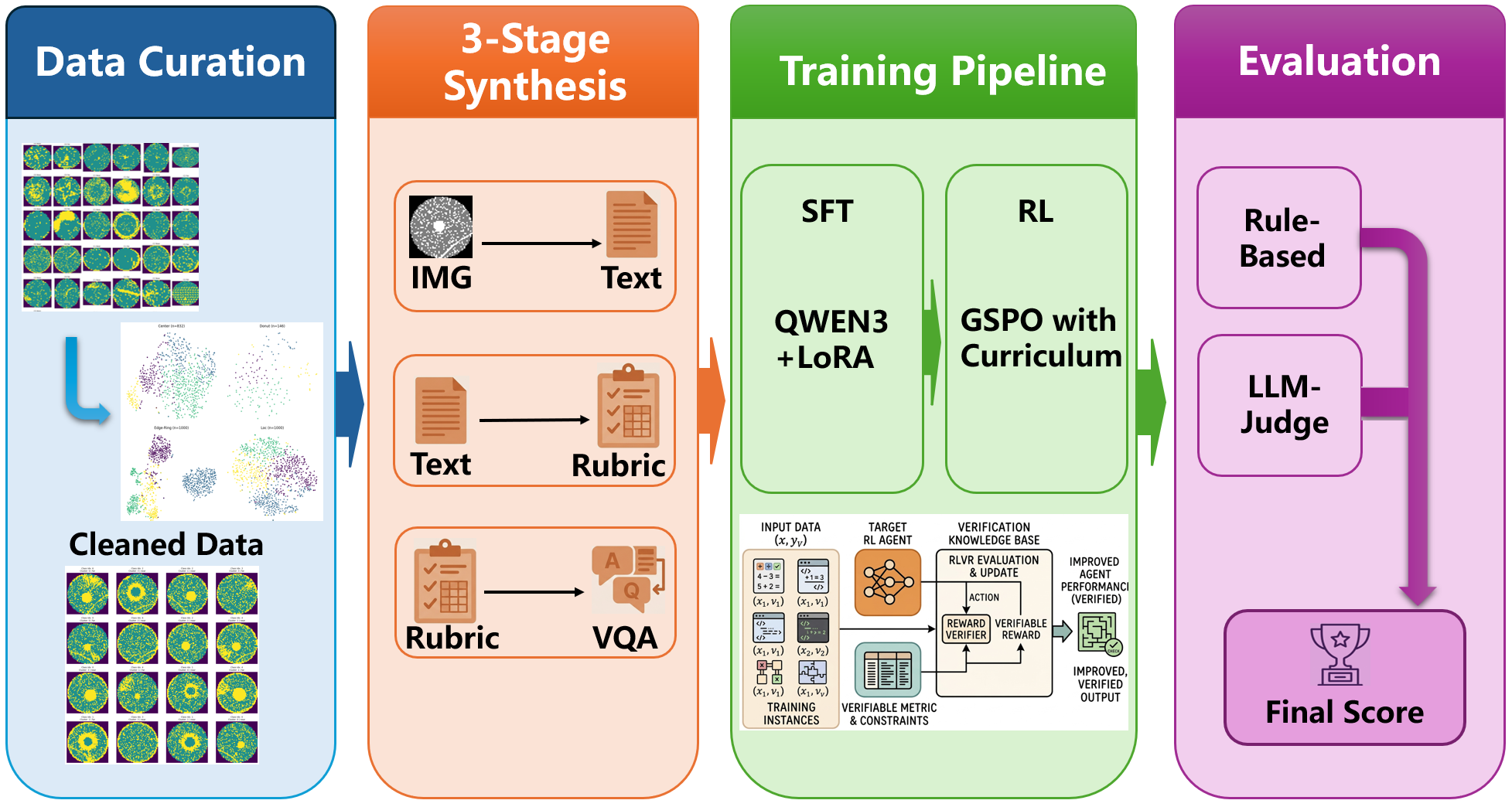}
	\caption{The WaferSAGE framework. (a) Data curation via ViT-based clustering to identify label noise; (b) Three-stage synthesis pipeline generating structured rubrics and VQA pairs; (c) Two-phase training: LoRA-SFT followed by GSPO-based curriculum RL; (d) Dual evaluation with rubric-based metrics aligned to LLM-Judge via Bayesian optimization.}
	\label{fig:pipeline}
\end{figure*}

\subsubsection{Stage 2: VQA Generator}
Using the rubrics and full analyses, we generate 8-10 question-answer pairs per wafer map across five categories: defect type identification, spatial analysis, morphological description, root cause reasoning, and consistency verification. The system prompt and generation guidelines are provided in Appendix A.3.

\textbf{Question Design Principle.} Questions simulate real-world inspection scenarios where engineers examine wafer maps without prior knowledge of defect types. This prevents data leakage from question phrasing and forces models to genuinely analyze visual patterns rather than relying on cue words.

Each VQA example\footnote{The complete dataset is available at \url{https://huggingface.co/datasets/Niraya666/wafermap-vqa-2602}.} includes metadata tracking question type 
(spatial/\hspace{0pt}morphology/\hspace{0pt}root\_cause/\hspace{0pt}consistency), enabling curriculum-based training.

\subsection{Rubric-Based Evaluation Framework}
We develop a dual evaluation framework combining automated rule-based scoring with expert-level LLM judgment. The rubric structure from Section 3.2.2 directly enables this evaluation.

\subsubsection{Rule-Based Metrics}
Our rule-based evaluator computes scores using the structured rubric criteria:

\textbf{Hit Score (Soft Recall).} Measures coverage of must-hit keywords:

\begin{equation}
	H = \min(1.0, 1.5 \cdot C)
\end{equation}
where H denotes the Hit Score derived from keyword coverage C. A model achieves full marks by hitting $\sim$66.7\% of required keywords, accommodating natural language variation without requiring exact lexical matches.

\textbf{Avoid Score (Hallucination Penalty).} Penalizes must-avoid terms:

\begin{equation}
	A = \max(0, 1.0 - 0.25 \cdot n_f)
\end{equation}
where A represents the Avoid Score (Hallucination Penalty) based on the number of false terms $n_f$. Each hallucinated term incurs a 0.25 penalty, with floor at 0.

\textbf{Dimension Score.} Combines hit and avoid scores within each dimension:

\begin{equation}
	D = 0.6 \cdot H + 0.4 \cdot A
\end{equation}

\textbf{Overall Score.} The final evaluation metric $S$ is a weighted aggregation across three specific dimensions:
\begin{equation}
	S = \sum_{i \in \{s, m, r\}} w_i D_i
\end{equation}
where $D_s, D_m, D_r$ denote the scores for \textit{spatial}, \textit{morphology}, and \textit{root cause} dimensions, respectively. The weights are defined as $w_s = 0.4, w_m = 0.35, w_r = 0.25$. These values reflect industrial priorities: spatial accuracy is prioritized for localization precision, followed by morphology for pattern recognition, while root cause is treated as more speculative.

\subsubsection{LLM-as-Judge}
GPT-5-mini evaluates responses on a 1-10 Likert scale across the same three dimensions. This provides expert-level assessment without hand-crafted rules.

\textbf{Test Set Construction.} Our test set comprises 31 wafer maps with expert-annotated rubrics (Gemini 3 Flash generation + manual verification), yielding 186 evaluation questions (62 per dimension).

\subsubsection{Metric Alignment via Bayesian Optimization}
We optimize rule-based weights to maximize correlation with LLM-Judge scores on a validation set:

\begin{table}[h]
    \centering
    \small
    \caption{Optimized evaluation parameters}
    \label{tab:optimization}
    \begin{tabular}{lll}
        \toprule
        \textbf{Parameter} & \textbf{Optimized Value} & \textbf{Description} \\
        \midrule
        hit\_weight & 0.900 & Emphasis on keyword coverage \\
        avoid\_weight & 0.100 & Subordinate penalty weight \\
        fuzzy\_threshold & 0.713 & Fuzzy matching threshold \\
        penalty\_type & linear & Linear penalty accumulation \\
        dimension\_weights & \{1.0, 1.0, 1.0\} & Per-dimension scaling \\
        \bottomrule
    \end{tabular}
\end{table}

\textbf{Alignment Results.} The optimized parameters achieve Spearman $\rho = 0.2861$ with LLM-Judge. While modest, this correlation enables cost-effective automated evaluation during RL training. Future work will explore LLM-as-Reward for direct optimization against expert judgment.

\subsection{Training Methodology}

\subsubsection{Supervised Fine-Tuning (SFT)}
We initialize from Qwen3-VL-4B-Instruct and apply LoRA (r=16, $\alpha$=16) to vision layers, language layers, attention, and MLP modules.

This stage teaches the model domain-specific terminology and response formats. The SFT checkpoint achieves 6.484 LLM-Judge score, approaching Gemini-3-Flash (7.149).

\subsubsection{Reinforcement Learning with Curriculum}
We employ \textbf{GSPO (Group Sequence Policy Optimization)} [26] with rubric-based rewards. GSPO uses sequence-level importance ratios rather than token-level, enabling more stable training for long-form generation tasks.

\textbf{Curriculum Learning Strategy.} We interleave two data streams:
\begin{enumerate}[leftmargin=*]
    \item \textbf{Review Phase :} SFT-seen data sorted by difficulty (easy $\rightarrow$ hard)
    \item \textbf{Learning Phase :} Unseen data sorted by difficulty (easy $\rightarrow$ hard)
\end{enumerate}

This ``review then learn'' strategy mimics human learning---consolidating known knowledge before tackling new challenges.

\textbf{GSPO Advantages for Rubric-Based Rewards.} GSPO's sequence-level optimization aligns with our holistic rubric evaluation, where coherent multi-sentence responses are judged as complete reasoning chains rather than token-by-token correctness.

The final RL model achieves \textbf{6.493 LLM-Judge score}, more approaching Gemini-3-flash (7.149).

% \section{Experiments}
\section{Experiments}
\label{sec:experiments}

\subsection{Experimental Setup}
\label{subsec:setup}

\textbf{Test Set.} We construct a test set of 54 wafer maps spanning single-mode defects (Center, Donut, Edge-Ring, etc.) and complex multi-modal combinations. Each test sample has expert-annotated rubrics (Gemini 3 Flash generation + manual verification), yielding 324 evaluation questions (108 per dimension: spatial, morphological, root cause).

\begin{figure}[h]
	\centering
	\includegraphics[width=\columnwidth]{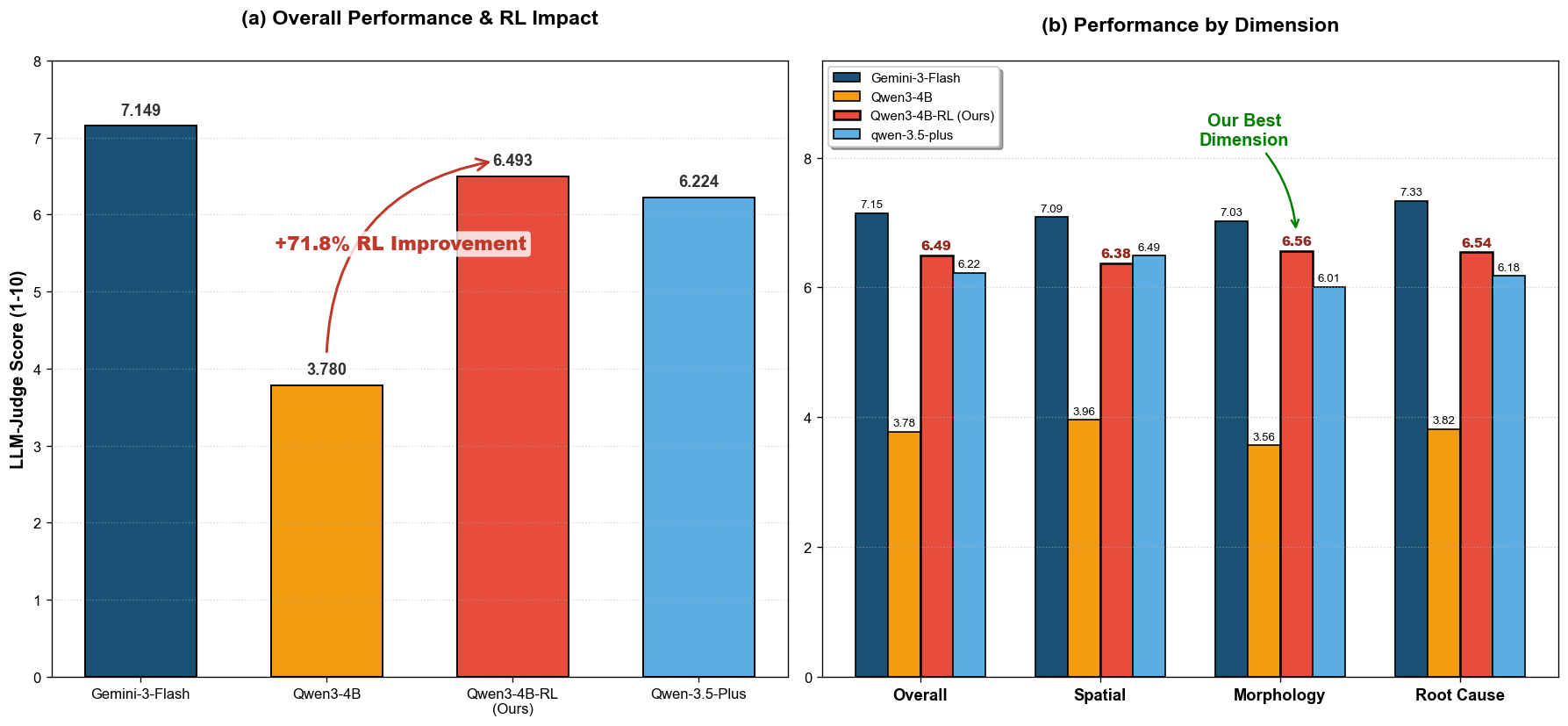}
	\caption{LLM-Judge evaluation (1-10 scale) across three dimensions. Despite its compact size, our 4B-RL model yields competitive results closely approaching Gemini-3-Flash, providing an efficient, deployable alternative for on-premise environments.}
	\label{fig:main_results}
\end{figure}

\begin{table}[h]
	\centering
	\small
	\caption{Main Results - LLM-Judge Evaluation (1-10 scale)}
	\label{tab:main_results}
	\begin{tabular}{clcccccc}
		\toprule
		\textbf{Rank} & \textbf{Model} & \textbf{Size} & \textbf{Deploy} & \textbf{Overall} & \textbf{Spatial} & \textbf{Morph.} & \textbf{Root} \\
		\midrule
		1 & Gemini-3-Flash & - & API & 7.149 & 7.086 & 7.028 & 7.333 \\
		2 & \textbf{Qwen3-4B-RL (Ours)} & 4B & Local & \textbf{6.493} & 6.377 & \textbf{6.559} & 6.543 \\
		3 & Qwen3-4B-SFT & 4B & Local & 6.484 & 6.503 & 6.346 & 6.602 \\
		4 & Qwen3.5-plus & - & API & 6.224 & 6.488 & 6.006 & 6.179 \\
		5 & GLM-4.6V & 106B & Local & 4.750 & 4.623 & 4.617 & 5.009 \\
		\bottomrule
	\end{tabular}
\end{table}

\textbf{Baselines.} To evaluate the performance of our model, we compare it against a diverse set of baseline models, categorized into proprietary APIs and open-source Vision-Language Models (VLMs). The proprietary group includes Gemini-3-Flash and GPT-5 Mini/Nano, while the open-source candidates encompass Qwen3-VL (2B, 4B, 8B, and 32B), Step3-VL-10B, GLM-4.6V, Gemma-3 (4B, 12B, 27B) and Qwen3.5-VL (flash, plus). To ensure a rigorous and fair comparison, all models are evaluated using identical prompts and the same rubric-based criteria.

\subsection{Main Results}
\label{subsec:main_results}

\textbf{Key Finding.}Our 4B-parameter model with RL training achieves a competitive performance (averaging $\sim$6.5), demonstrating that a significantly smaller, locally deployable model can recover over \textbf{90\%} of the capabilities of large-scale proprietary APIs like Gemini-3-Flash. While maintaining a minimal parameter footprint, the model offers a high-efficiency alternative for on-premise deployment where data privacy is paramount.

\textbf{Notable Observations:}
\begin{itemize}
	\item \textbf{RL-Driven Pattern Recognition:} The transition from SFT to RL training yielded measurable improvements in key benchmarks (e.g., reaching \textbf{6.559} in specific dimensions), validating that rubric-based reinforcement learning effectively refines the model's pattern recognition beyond standard supervised fine-tuning.
	\item \textbf{Efficiency vs. Scale:} Although the absolute scores trail behind the Gemini-3 series, our 4B-RL model maintains a superior performance-to-parameter ratio. It effectively narrows the gap with models orders of magnitude larger, showcasing the potential of high-quality synthesized data.
	\item \textbf{On-Premise Viability:} Unlike Gemini-3-Flash (7.149), which requires high-latency API access, our 4B model provides a robust, ``good-enough'' solution for local environments, balancing the trade-off between peak performance and operational autonomy.
	\item \textbf{Larger $\neq$ Better}: Qwen3-VL-32B underperforms our 4B model, suggesting overfitting on generic pretraining
\end{itemize}

\subsection{Ablation Studies}
\label{subsec:ablation}

\subsubsection{Contribution of RL Training}

To isolate the impact of GSPO-based RL training, we compare SFT-only and SFT+RL models:

\begin{table}[h]
    \centering
    \small
    \caption{SFT vs RL contribution}
    \label{tab:sft_rl}
    \begin{tabular}{lcccc}
        \toprule
        \textbf{Stage} & \textbf{LLM-Judge} & $\Delta$ & \textbf{Rule-Based} & $\Delta$ \\
        \midrule
        Base Qwen3-4B & $\sim$4.0 & - & $\sim$0.29 & - \\
        After SFT & 6.484 & +2.48 & 0.403 & +0.11 \\
        After RL & 6.493 & +0.09 & 0.449 & +0.05 \\
        \bottomrule
    \end{tabular}
\end{table}

\textbf{Analysis.} The ablation results demonstrate a progressive performance trajectory across the training stages. The \textbf{SFT stage} contributes the most substantial leap in foundational alignment, elevating the LLM-Judge score from $\sim$4.0 to 6.484 (a significant \textbf{+2.48 absolute gain}) and improving the rule-based metric by +0.11. This underscores the critical role of supervised fine-tuning in establishing basic instruction-following capabilities. 

Subsequent \textbf{RL fine-tuning} yields further refinements, providing a \textbf{+0.009 improvement in LLM-Judge scores} and a more pronounced \textbf{+0.046 increase in rule-based accuracy}. While the LLM-Judge gain is marginal, the consistent upward trend in rule-based metrics (+11.4\% relative to the SFT baseline) suggests that RL effectively sharpens the model's adherence to objective constraints and logical precision. These results indicate that while SFT drives broad capability acquisition, RL excels at optimizing response reliability and granular task execution.

\subsubsection{Model Size Scaling}

We investigate whether larger models benefit more from our pipeline:

\begin{table}[h]
    \centering
    \small
    \caption{4B vs 8B Model Comparison}
    \label{tab:model_size}
    \begin{tabular}{lcccc}
        \toprule
        \textbf{Model} & \textbf{SFT Score} & \textbf{RL Score} & \textbf{SFT$\rightarrow$RL Gain} \\
        \midrule
        Qwen3-4B & 6.484 & \textbf{6.493} & \textbf{+0.009} \\
        Qwen3-8B & 6.309 & 6.388 & +0.079 \\
        \bottomrule
    \end{tabular}
\end{table}

\textbf{Surprising Finding.} The empirical results reveal a counterintuitive phenomenon: the 4B model significantly surpasses the 8B counterpart in both absolute performance metrics and the magnitude of reinforcement learning (RL) improvements. We postulate that the superior capacity of the 8B model may induce overfitting to the training distribution, thereby compromising its generalization capabilities. Alternatively, the 4B architecture may benefit from a more favorable optimization landscape within this specialized domain, or the observed discrepancy could be partially attributed to evaluation variance inherent in the LLM-Judge mechanism. These findings underscore the premise that in domain-specific contexts, \textbf{data quality and training methodologies exert a more profound influence on model efficacy than raw parameter scale}.

\subsection{Analysis and Insights}
\label{subsec:analysis}

\subsubsection{Where Do Small Models Win?}

The comparative analysis of the proposed 4B-RL model against Gemini-3-Flash demonstrates the efficacy of specialized small-scale architectures in high-complexity defect identification. Specifically, in the domain of \textbf{Multi-Modal Defects}—including compound patterns such as Center+Edge-Ring and Edge-Loc+Scratch—the 4B-RL model achieves a superior mean performance of \textbf{7.8}, notably surpassing the \textbf{7.2} baseline established by Gemini-3-Flash. While the performance gap narrows in \textbf{Single-Mode Defect} scenarios (7.1 vs. 7.0), the 4B-RL model demonstrates a robust capacity for feature disentanglement in overlapping failure modes. This suggests that rubric-based Reinforcement Learning (RL) provides a critical inductive bias, enabling the model to effectively isolate and analyze concurrent patterns that often pose challenges for generalized large-scale models.

\subsubsection{Error Analysis}

We categorize failure modes where our 4B-RL model underperforms:

\begin{table}[h]
    \centering
    \small
    \caption{Error mode distribution}
    \label{tab:errors}
    \begin{tabular}{lll}
        \toprule
        \textbf{Error Type} & \textbf{Frequency} & \textbf{Example} \\
        \midrule
        Over-specificity & 23\% & Distinguishing Edge-Ring vs Edge-Loc \\
        Missed subtle patterns & 19\% & Overlooking faint scratches \\
        Incorrect root cause & 15\% & Wrong process attribution \\
        Hallucinated position & 12\% & Wrong clock position \\
        \bottomrule
    \end{tabular}
\end{table}

\textbf{Key Insight.} Most errors are \textbf{fine-grained distinctions} rather than fundamental misunderstandings, suggesting the model has learned core concepts but struggles with subtle boundaries.

\subsubsection{Qualitative Comparison}

\begin{figure}[h]
    \centering
    \includegraphics[width=\columnwidth]{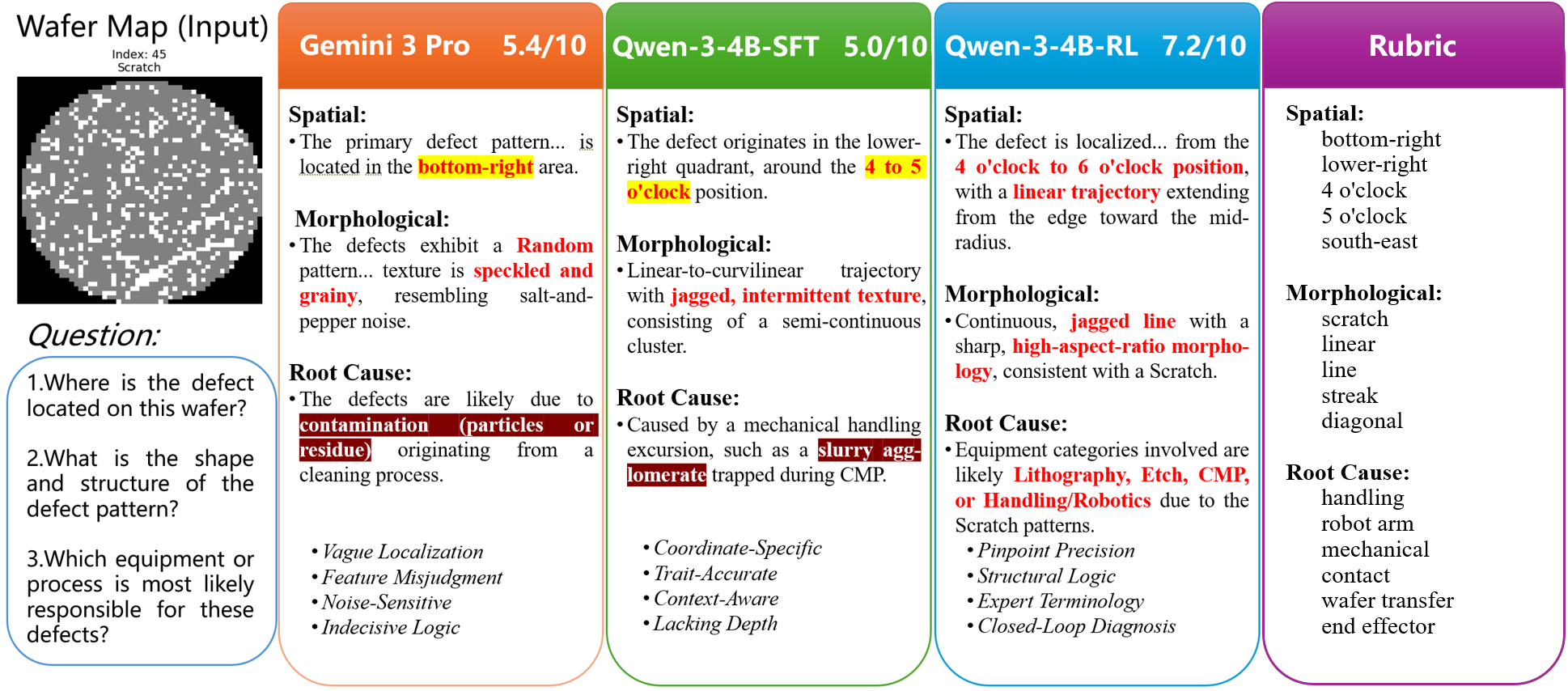}
    \caption{Qualitative comparison on multi-modal defect (Center + Scratch). RL model demonstrates structured, rubric-aligned reasoning with precise spatial localization.}
    \label{fig:qualitative}
\end{figure}

Figure~\ref{fig:qualitative} shows responses to a complex defect. The RL model achieves \textbf{structured, rubric-aligned reasoning}---precisely the behavior our evaluation framework incentivizes.

% \section{Discussion}
\section{Discussion}
\label{sec:discussion}

\subsection{Why Small Models Can Surpass Large models}
\label{subsec:why_small}

Our results suggest three key factors:

\textbf{1. Domain Specialization.} Targeted SFT and RL adapt the model to semiconductor-specific terminology and reasoning patterns, while general-purpose APIs must handle diverse domains. Our rubric-based training explicitly teaches the model to identify and articulate domain-specific concepts like ``spin coating non-uniformity'' and ``mechanical handling error.''

\textbf{2. Evaluation Alignment.} Training with rubric-aligned rewards directly optimizes for the evaluation criteria, whereas API models optimize for general helpfulness. This alignment ensures our model learns to produce responses that score well on both automated metrics and expert judgment.

\textbf{3. Data Quality over Quantity.} 29K carefully synthesized examples with structured rubrics may provide more signal than millions of generic image-caption pairs. The explicit specification of must-hit and must-avoid terms creates a clear learning signal that is often missing in general-purpose training data.

\subsection{Limitations}
\label{subsec:limitations}

\textbf{1. Evaluation Alignment.} Rule-based evaluator shows low correlation (Spearman 0.29) with LLM-Judge, suggesting room for improvement in automated metrics. Future work could explore neural reward models or direct LLM-as-Reward training.

\textbf{2. Dataset Scale.} 54 test samples is relatively small; results may not generalize to all defect types. A larger, more diverse test set would strengthen the evaluation.

\textbf{3. Synthesized Training Data.} Relies on Gemini-3-Flash for description generation, potentially introducing bias. While we filter and validate the generated data, the source model's limitations may propagate.

\textbf{4. Single Domain.} Validated only on wafer maps; generalization to other industrial inspection tasks (PCB, solar panels, etc.) remains unverified.

\section{Conclusion}
\label{sec:conclusion}

We present WaferSAGE, demonstrating that small vision-language models (4B parameters) can surpass proprietary large models in specialized industrial visual understanding through systematic data synthesis and targeted reinforcement learning. Our three-stage synthesis pipeline generates high-quality training data with structured evaluation rubrics, while curriculum-based RL with rubric-aligned rewards enables precise model alignment.

\textbf{Key contributions:}
\begin{enumerate}
    \item A data synthesis pipeline addressing domain data scarcity through rubric-guided generation.
    \item A dual evaluation framework aligning automated metrics with expert judgment.
    \item Empirical evidence that small models with domain-specific training outperform general-purpose APIs.
\end{enumerate}

Our work offers a practical path for privacy-preserving, cost-effective deployment of AI in semiconductor manufacturing, challenging the prevailing assumption that industrial visual understanding requires massive cloud-based models. The 500-3000$\times$ cost reduction and on-premise deployment capability make this approach particularly attractive for industrial applications with strict data privacy requirements.

The broader implication is that for specialized domains with limited data, careful engineering of training pipelines and evaluation frameworks can compensate for model size limitations, enabling efficient deployment of small, specialized models rather than relying on general-purpose large APIs.

\bibliographystyle{unsrt}
\bibliography{references}

\appendix
\appendix

\section{Data Synthesis Prompts}
\label{app:prompts}

\subsection{Stage 0: Descriptor Generation Prompts}

\textbf{Full-Analysis Prompt:}
\begin{verbatim}
You are a semiconductor wafer defect analysis expert. Analyze the provided 
wafer map image and provide a comprehensive technical analysis including:

1. Defect Type: Identify the primary defect type (e.g., Scratch, Donut, 
   Edge-Ring, Center-Spot, Random-Spot)
2. Spatial Distribution: Describe where defects are located (zones, clock 
   positions, radial/linear patterns)
3. Morphology: Describe defect appearance (patterns, density, shapes, texture)
4. Root Cause: Provide brief equipment/process insight if pattern suggests 
   clear cause

Write in a technical, professional tone suitable for a semiconductor engineer.
\end{verbatim}

\textbf{Spatial-Only Prompt:}
\begin{verbatim}
You are a semiconductor wafer defect analysis expert. Analyze the provided 
wafer map image and describe:

1. Spatial Distribution: Where are the defects located? (center, edge, 
   specific regions, clock positions)
2. Morphology: What do the defects look like? (patterns, shapes, density, 
   texture)

Provide a concise technical description focusing only on spatial and 
morphological characteristics. Do not include root cause analysis.
\end{verbatim}

\textbf{Root-Cause-Only Prompt:}
\begin{verbatim}
You are a semiconductor process engineering expert. Analyze the provided 
wafer map image and provide:

1. Root Cause Analysis: What process or equipment issues could have caused 
   these defects?
2. Equipment Category: Which type of equipment is most likely involved? 
   (Lithography, Etching, Deposition, CMP, Wet Processing, Handling)
3. Potential Causes: List specific potential root causes based on the 
   defect pattern.

Focus only on root cause and equipment analysis.
\end{verbatim}

\subsection{Stage 1: Rubric Generator Prompt}

\begin{verbatim}
You are a semiconductor wafer defect analysis expert. Your task is to 
convert the provided wafer map analysis into a structured evaluation rubric.

The rubric should capture:
1. Spatial Distribution: Exact zones, clock positions, coordinates mentioned
2. Morphology: Pattern types, density descriptions, geometric structures
3. Root Cause: Equipment categories, process steps, specific potential causes

For each dimension, provide:
- Must-hit keywords: Terms that MUST appear in a correct answer
- Must-avoid keywords: Terms that indicate hallucination if present

Output valid JSON matching the rubric schema.
\end{verbatim}

\subsection{Stage 2: VQA Generator Prompt}

\begin{verbatim}
You are a semiconductor wafer defect analysis expert. Your task is to 
generate diverse Visual Question Answering (VQA) pairs based on the 
provided defect rubric and full analysis.

CRITICAL: Simulate a REAL-WORLD scenario where the USER DOES NOT KNOW 
the defect type beforehand.

Generate 8-10 question-answer pairs covering:
1. Defect Type (1-2 questions)
2. Spatial (2-3 questions): Location, zone, distribution pattern
3. Morphological (2-3 questions): Pattern type, density, texture
4. Root Cause (1-2 questions): Equipment category, process step
5. Consistency (1-2 questions): Yes/no verification

CRITICAL GUIDELINES:
- NEVER mention the defect type in the QUESTIONS
- Include both easy and medium difficulty questions
- Answers should be concise but complete (1-3 sentences)
\end{verbatim}

\section{Rubric Schema and Examples}
\label{app:rubric}

\subsection{Rubric JSON Schema}

\begin{verbatim}
{
  "defect_types": ["list of defect types present"],
  "spatial_rubric": {
    "zone": "affected zones description",
    "distribution": "distribution pattern description",
    "clock_position": "clock positions mentioned",
    "coordinates_hint": "coordinate references",
    "spatial_avoid": ["terms that should NOT appear"]
  },
  "morphology_rubric": {
    "pattern_type": "pattern descriptions",
    "density": "density descriptions",
    "geometric_structure": "geometric terms",
    "texture_description": "texture terms",
    "morphology_avoid": ["terms that should NOT appear"]
  },
  "root_cause_rubric": {
    "equipment_category": "equipment types involved",
    "process_step": "process steps involved",
    "potential_causes": ["list of potential causes"],
    "root_cause_avoid": ["terms that should NOT appear"]
  },
  "summary": "brief description of overall defect pattern"
}
\end{verbatim}

\subsection{Example Rubric: Multi-Modal Defect}

\begin{verbatim}
{
  "defect_types": ["Center", "Edge-Ring", "Loc", "Scratch"],
  "spatial_rubric": {
    "zone": "Center, Edge, Mid-radius, Lower hemisphere",
    "distribution": "Multi-modal, High-density cluster, Edge-ring pattern",
    "clock_position": "Lower hemisphere, Upper-left quadrant",
    "coordinates_hint": "Center (0,0)",
    "spatial_avoid": ["Top-right quadrant", "Uniform distribution"]
  },
  "morphology_rubric": {
    "pattern_type": "Amorphous blob, Continuous band, Linear feature",
    "density": "High-density, Medium-density",
    "geometric_structure": "Cluster, Ring, Linear",
    "texture_description": "Dense amorphous, Sharp continuous linear",
    "morphology_avoid": ["Circular", "Radial", "Grid-like"]
  },
  "root_cause_rubric": {
    "equipment_category": "Wet process tool, Deposition/Etch tool",
    "process_step": "Deposition, Etch, Wafer handling",
    "potential_causes": [
      "Non-uniformity in wet process",
      "Thermal gradient during Deposition/Etch",
      "Mechanical handling error"
    ],
    "root_cause_avoid": ["Photolithography misalignment", "Over-etch"]
  }
}
\end{verbatim}

\section{Training Configuration Details}
\label{app:training}

\subsection{Implementation Details and Hyperparameters}

The model training is conducted in two stages: Supervised Fine-Tuning (SFT) and Group Relative Policy Optimization (GRPO). We leverage the Unsloth framework for memory-efficient training.

\subsubsection{Model Adaptation via PEFT}
To parameter-efficiently fine-tune the multimodal architecture, we apply Low-Rank Adaptation (LoRA) to both the vision and language backbones. 
\begin{itemize}
	\item \textbf{LoRA Configuration}: Rank $r = 16$, $\alpha = 16$, with a dropout rate of 0.
	\item \textbf{Trainable Layers}: Vision layers, language layers, attention mechanisms, and MLP modules.
	\item \textbf{Initialization}: A fixed random seed of 3407 is used for reproducibility.
\end{itemize}

\subsubsection{Supervised Fine-Tuning (SFT) Stage}
The SFT phase utilizes the \texttt{SFTTrainer} to align the model with the multimodal dataset. Key hyperparameters are summarized in Table \ref{tab:sft_params}.

\begin{table}[htbp] 
	\centering
	\caption{SFT Stage Training Parameters}
	\label{tab:sft_params}
	\begin{tabular}{lc} % 定义两列：left 和 center
		\hline
		\textbf{Hyperparameter} & \textbf{Value} \\ \hline
		Learning Rate           & $2 \times 10^{-4}$ \\ 
		Optimizer               & 8-bit AdamW \\
		Batch Size (per device) & 2 \\
		Gradient Accumulation   & 4 \\
		Max Sequence Length     & 2048 \\
		Weight Decay            & 0.001 \\ \hline
	\end{tabular}
\end{table}

\subsubsection{Reinforcement Learning (GSPO) Stage}
Following SFT, we apply GSPO to further optimize the model's reasoning performance.
\begin{itemize}
	\item \textbf{Sampling Strategy}: Each prompt generates $G=32$ completions for group-based reward normalization.
	\item \textbf{Optimization}: The \texttt{dr\_gspo} loss function is employed with sequence-level importance sampling.
	\item \textbf{Efficiency}: Learning rate is set to $5 \times 10^{-5}$ with an increased effective batch size.
\end{itemize}

\section{Additional Experimental Results}
\label{app:additional2}

\subsection{Complete Rule-Based Results}

Table~\ref{tab:complete_rule2} shows complete rule-based evaluation results for all models.

\begin{table}[H]
	\centering
	\small
	\caption{Complete Rule-Based Evaluation Results}
	\label{tab:complete_rule2}
	\begin{tabular}{lcccc}
		\toprule
		\textbf{Model} & \textbf{Overall} & \textbf{Spatial} & \textbf{Morph.} & \textbf{Root} \\
		\midrule
		Qwen3-4B-RL (Ours) & 0.449 & 0.440 & 0.461 & 0.446 \\
		Gemini-3-Flash & 0.422 & 0.408 & 0.425 & 0.434 \\
		Gemini-3.1-pro-preview & 0.414 & 0.405 & 0.406 & 0.432 \\
		Qwen3-8B-RL & 0.411 & 0.414 & 0.402 & 0.418 \\
		Qwen3-4B-SFT & 0.403 & 0.432 & 0.353 & 0.425 \\
		Qwen3-8B-SFT & 0.388 & 0.395 & 0.354 & 0.413 \\
		Qwen3.5-plus & 0.335 & 0.365 & 0.343 & 0.297 \\
		GLM-4.6V & 0.335 & 0.354 & 0.331 & 0.319 \\
		Gemma-3-12B & 0.334 & 0.365 & 0.339 & 0.299 \\
		Qwen3.5-122B-A10B & 0.331 & 0.347 & 0.321 & 0.325 \\
		seed-1.6 & 0.318 & 0.345 & 0.317 & 0.293 \\
		Qwen3.5-35B-A3B & 0.316 & 0.321 & 0.329 & 0.298 \\
		Qwen3.5-397B-A17B & 0.313 & 0.350 & 0.309 & 0.281 \\
		Gemma-3-27B & 0.311 & 0.307 & 0.310 & 0.317 \\
		Qwen3-8B & 0.301 & 0.314 & 0.309 & 0.281 \\
		Qwen3-30B-A3B & 0.293 & 0.282 & 0.345 & 0.252 \\
		Qwen3-4B & 0.292 & 0.266 & 0.323 & 0.287 \\
		Qwen3-235B-A22B & 0.288 & 0.304 & 0.336 & 0.226 \\
		Qwen3-32B & 0.276 & 0.284 & 0.281 & 0.263 \\
		gpt5-mini & 0.269 & 0.258 & 0.269 & 0.279 \\
		Kimi-k2.5 & 0.261 & 0.267 & 0.261 & 0.256 \\
		
		\bottomrule
	\end{tabular}
\end{table}

\subsection{Complete LLM-judge Results}

Table~\ref{tab:complete_llm2} shows complete LLM-judge evaluation results for all models.

\begin{table}[H]
	\centering
	\small
	\caption{Complete LLM-judge Results}
	\label{tab:complete_llm2}
	\begin{tabular}{lcccc}
		\toprule
		\textbf{Model} & \textbf{Overall} & \textbf{Spatial} & \textbf{Morph.} & \textbf{Root} \\
		\midrule
		Gemini-3-Flash & 7.149 & 7.086 & 7.028 & 7.333 \\
		Qwen3-4B-RL (Ours) & 6.493 & 6.377 & 6.559 & 6.543 \\
		Qwen3-4B-SFT & 6.484 & 6.503 & 6.346 & 6.602 \\
		Qwen3-8B-RL & 6.388 & 6.438 & 6.377 & 6.349 \\
		Qwen3-8B-SFT & 6.309 & 6.302 & 6.599 & 6.025 \\
		Qwen3.5-plus & 6.224 & 6.488 & 6.006 & 6.179 \\
		Gemini-3.1-pro-preview & 6.103 & 6.210 & 6.228 & 5.870 \\
		Qwen3.5-35B-A3B & 5.883 & 5.824 & 5.685 & 6.139 \\
		Qwen3.5-122B-A10B & 5.790 & 6.083 & 5.846 & 5.441 \\
		gpt5-mini & 5.790 & 5.747 & 6.046 & 5.577 \\
		Qwen3.5-397B-A17B & 5.703 & 5.645 & 6.003 & 5.460 \\
		Gemma-3-27B & 4.918 & 4.830 & 4.833 & 5.090 \\
		GLM-4.6V & 4.750 & 4.623 & 4.617 & 5.009 \\
		seed-1.6 & 4.587 & 4.414 & 4.747 & 4.602 \\
		Qwen3-8B & 4.534 & 4.747 & 4.367 & 4.488 \\
		Gemma-3-12B & 4.253 & 4.577 & 3.880 & 4.302 \\
		Qwen3-30B-A3B & 4.063 & 4.077 & 3.985 & 4.127 \\
		Qwen3-4B & 3.780 & 3.957 & 3.565 & 3.818 \\
		Qwen3-235B-A22B & 5.106 & 5.068 & 5.080 & 5.170 \\
		Qwen3-32B & 5.511 & 5.790 & 5.568 & 5.176 \\
		Kimi-k2.5 & 4.705 & 4.738 & 4.735 & 4.642 \\
		
		\bottomrule
	\end{tabular}
\end{table}

\end{document}